\documentclass{article} \usepackage{iclr2027_conference,times}

\usepackage{amsmath,amsfonts,bm}

\def\eqref#1{equation~\ref{#1}}

\def\1{\bm{1}}

\DeclareMathAlphabet{\mathsfit}{\encodingdefault}{\sfdefault}{m}{sl}
\SetMathAlphabet{\mathsfit}{bold}{\encodingdefault}{\sfdefault}{bx}{n}

\usepackage{hyperref}
\usepackage{url}
\usepackage{booktabs,multirow,graphicx,longtable}
\usepackage{algorithm,algpseudocode}

\title{Adaptive Consistency Graph for Long-Horizon Agents}

\author{Jiecong Wang \\
Beihang University \\
\texttt{jcwang@buaa.edu.cn}
\And
Hao Pneg \\
Beihang University \\
\texttt{penghao@buaa.edu.cn}
\And
Zhanyi Wang \\
QI-ANXIN GROUP \\
\texttt{wangzhanyi@qianxin.com}
}

\newcommand{\rate}[2]{\ensuremath{#1\,\mathbin{\pm}_{\scriptscriptstyle #2}}}
\newcommand{\bfrate}[2]{\ensuremath{\bm{#1\,\mathbin{\pm}_{\scriptscriptstyle #2}}}}

\iclrfinalcopy
\begin{document}

\maketitle
\fancyhead{}
\renewcommand{\headrulewidth}{0pt}

\begin{abstract}
Large language model agents can often make reasonable local decisions on short tasks, yet their performance degrades when success requires long sequences of dependent actions and tool calls. During execution, task requirements, historical evidence, and the current execution state may gradually become disconnected, so later decisions can drift from the original objective. We study this problem by introducing the Adaptive Consistency Graph (ACG) for long-horizon execution. ACG incrementally organizes execution evidence and its provenance in a persistent graph, then constructs a temporary requirement-centered view for each decision under a bounded context budget. Rather than replacing the base agent's planner or tool executor, ACG provides a structured and traceable context view for each decision. In the matched evaluation, ACG improves GPT-5.6-luna's average success from 44.5\% with ReAct to 50.2\%, with the largest gain on BrowseComp-Plus (73.5\% versus 62.4\%). We further analyze trajectory structure and inference cost to characterize this improvement.
Our code is available at \url{https://github.com/yunsaijc/Adaptive-Consistency-Graph}.
\end{abstract}

\section{Introduction}
\label{sec:introduction}
Large language models are increasingly used as agents that pursue goals
through sequences of actions and environmental feedback. Work on grounded
robotic control, feedback-driven planning, and open-ended skill acquisition
demonstrates how language models can participate in extended interaction
rather than produce a single response
\citep{ichter2023saycan,huang2023innermonologue,wang2024voyager}.
In these settings, later decisions depend on information accumulated during
earlier execution. An agent must retain relevant task requirements,
interpret previous observations, and relate them to the situation it
currently faces.

Advances in intermediate reasoning and search have strengthened the ability
of language models to solve complex problems
\citep{wei2022cot,yao2023tot,zhou2024lats}. Their decisions nevertheless
depend on the information available at each step. Recurrent, compressed,
and retrieval-based memory architectures have expanded access to earlier
context \citep{dai2019transformerxl,rae2020compressive,wu2022memorizing},
while studies of long-context behavior show that information can remain
difficult to use even when it fits within the context window
\citep{liu2024lost}. For interactive agents, these findings motivate
studying how execution history is organized and presented alongside
improvements in reasoning and context capacity.

Agent memory research provides several approaches to managing accumulated
information. Generative Agents combines memory retrieval, reflection, and
planning \citep{park2023generative}; Reflexion and ExpeL use feedback and
experience to inform subsequent attempts or tasks
\citep{shinn2023reflexion,zhao2024expel}. HiAgent organizes working memory
around subgoals and supports access to earlier interaction details
\citep{hu2025hiagent}. These approaches highlight the importance of
selecting and abstracting experience. They also motivate a more specific
question for ongoing execution: how should information associated with
different events and task requirements be brought together when
constructing the context for the next decision?

Retrieval-based methods offer another foundation for constructing useful
context. Retrieval-augmented generation and passage fusion demonstrate how
external evidence can support language generation
\citep{lewis2020rag,izacard2021fid}, while Self-RAG incorporates decisions
about retrieval and evidence use into generation \citep{asai2024selfrag}.
LongMem extends access to past context through an external memory mechanism
\citep{wang2023longmem}. HippoRAG and its successor further explore
graph-based associative retrieval and the integration of information
across documents \citep{gutierrez2024hipporag,gutierrez2025hipporag2}.
These results motivate relational memory representations, while their
application to evolving task execution requires attention to how new
observations are incorporated and how relevant execution context is
selected.

We focus on context construction during an ongoing task. A useful execution
context should connect relevant historical evidence with the current
decision, preserve enough surrounding information to interpret that
evidence, and allocate detail according to the available context budget.
For example, retrieving a relevant observation may be insufficient when
its interpretation depends on the event that produced it or on related
observations elsewhere in the trajectory. This motivates studying
execution memory as an evolving relational structure whose organization
and readout are jointly designed for the agent's immediate needs.

We introduce \textbf{Adaptive Consistency Graph (ACG)}, an approach to
organizing and accessing execution information for long-horizon agents.
ACG stores task-relevant evidence as source-linked memory units and
retains their event occurrences and relations. At each decision, the task
and the latest interaction seed retrieval; the selected records are then
organized into a temporary requirement-centered view and rendered under
the remaining context budget. Adaptation therefore concerns the records and
detail exposed to the base agent at each decision, while the underlying
memory remains available for later decisions. ACG supplies this view
without replacing the base planner or tool executor.

Our empirical study examines task success and inference cost across
constrained planning, corpus-based information retrieval, and repository-level
code repair. Baseline analyses describe how outcomes vary with task structure,
observed execution length, and model-call expenditure.
These observations motivate the evaluation of ACG, but do not identify
inconsistent state as the cause of failure. Establishing an improvement in
task success and cost is distinct from attributing that improvement to an
individual memory mechanism.

Our contributions are:
\begin{list}{\textbullet}{\setlength{\leftmargin}{0pt}\setlength{\itemindent}{0pt}\setlength{\labelsep}{0.5em}\setlength{\itemsep}{1pt}\setlength{\topsep}{3pt}}
\item \textbf{Persistent execution memory.} ACG links evidence units to
their source events and related records, preserving details for later retrieval.
\item \textbf{Adaptive decision views.} Task- and event-seeded retrieval
selects evidence for requirement-qualified views under a token budget,
without overwriting the persistent memory.
\item \textbf{Empirical evaluation.} Across three benchmarks and two models,
we compare success, execution behavior, and inference cost against five baselines.
\end{list}

\section{Related Work}
\label{sec:related-work}
\subsection{Long-Horizon Planning and Interactive Agents}
Language-model agents extend single-turn generation by alternating reasoning
with actions whose consequences are observed from an environment. ReAct makes
this coupling explicit, while SayCan grounds language-model proposals in
robotic affordances and Inner Monologue feeds environmental feedback back into
planning~\citep{yao2023react,ichter2023saycan,huang2023innermonologue}.
Search-based methods such as Tree of Thoughts and Language Agent Tree Search
expand and evaluate alternative reasoning or action paths before committing
to a decision~\citep{yao2023tot,zhou2024lats}. Voyager instead accumulates
reusable skills during open-ended interaction~\citep{wang2024voyager}.
Together, these studies improve action selection, feedback use, or planning
search. A complementary challenge arises after execution has accumulated many
steps: the agent must keep the original requirements, prior evidence, and
latest state mutually consistent while deciding what context to expose next.
ACG addresses this execution-time context problem and leaves the base planner
and action interface unchanged.

\subsection{Memory and Context Management for Long-Horizon Agents}
Agent-memory work gives language-model systems access to information beyond
the current prompt, but differs in what is stored and how it is reused.
Generative Agents retrieves and reflects over a memory stream to support
ongoing behavior~\citep{park2023generative}; Reflexion stores verbal feedback
between attempts~\citep{shinn2023reflexion}; and ExpeL distills reusable
lessons from successful and failed trajectories~\citep{zhao2024expel}.
Hierarchical working-memory methods organize intermediate information at
multiple levels to support long tasks~\citep{hu2025hiagent}. These systems
primarily emphasize persistence, reflection, or reusable abstractions. A
generated summary can reduce prompt length, but it may discard the local event
and source needed to reconstruct why a fact was recorded; when a later decision
needs that detail, the agent must recover it from an already-compressed
representation. ACG uses the current execution record as its primary evidence archive: it
preserves event-level evidence and provenance, then constructs a temporary
decision view without replacing the underlying record with a generated
summary.

Long-context systems address the finite input window through segment-level
compression, recurrent memory, or retrieval from an external store.
Transformer-XL reuses hidden states across segments; Compressive Transformers
retain compressed representations of older activations; Memorizing Transformers
and LongMem augment the model with retrieval-accessible memory
~\citep{dai2019transformerxl,rae2020compressive,wu2022memorizing,wang2023longmem}.
However, access to more history does not ensure reliable use of the relevant
parts: Lost in the Middle shows that information use can depend strongly on
position within a long context~\citep{liu2024lost}. Recent agent methods adapt
context handling to execution: ACON optimizes history-compression guidelines,
COMPASS separates action execution from evolving context management, and TDP
decouples planning from execution through task-specific plans
~\citep{kang2025acon,wan2025compass,li2026tdp}. These methods motivate the
need for bounded views, while ACG additionally keeps omitted evidence in a
source-addressable execution record for later retrieval.

\subsection{Relational Retrieval and Graph Memory}
Retrieval-augmented generation selects external passages to condition a
language model~\citep{lewis2020rag}; Fusion-in-Decoder studies how retrieved
passages can be jointly used during generation, and Self-RAG makes retrieval
and critique part of the generation process~\citep{izacard2021fid,asai2024selfrag}.
Graph-based memory adds explicit relations to retrieval: HippoRAG and
HippoRAG~2 use associative structures to connect entities and support
multi-hop access to knowledge~\citep{gutierrez2024hipporag,gutierrez2025hipporag2}.
ACG shares the premise that relations can help locate relevant information,
but its graph represents a task's evolving execution rather than a static
knowledge collection. Its nodes retain links to the events and tool outputs
from which they were derived, and retrieval is conditioned on the current
requirements and state. Thus, the graph organizes evidence for execution; it
does not itself assert that retrieved content is true or that a requirement
has been satisfied.

\section{Adaptive Consistency Graph}
\label{sec:method}
\begin{figure*}[t]
\centering
\includegraphics[width=0.96\textwidth]{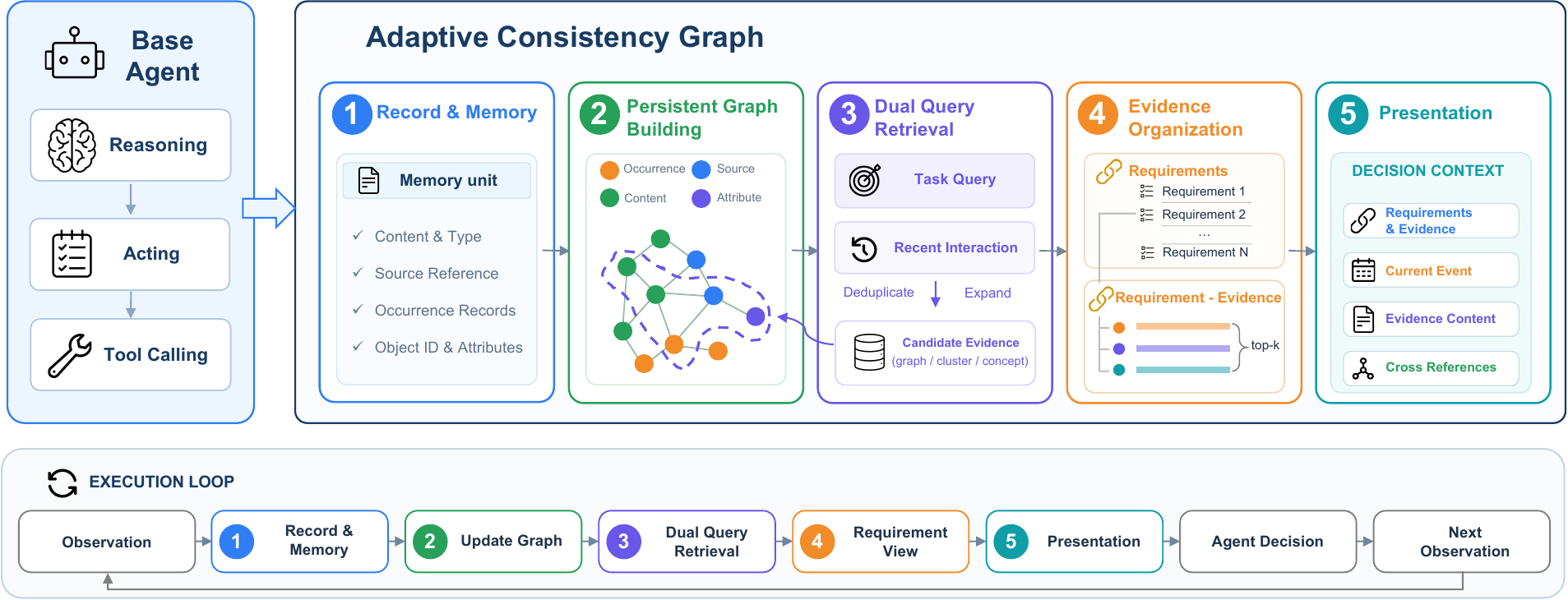}
\caption{Overview of Adaptive Consistency Graph. ACG retains a source-grounded execution graph and produces a budgeted, requirement-centered view for the base agent at each decision. The view is temporary; the underlying event record remains available for subsequent steps.}
\label{fig:acg-overview}
\end{figure*}
\subsection{Problem Formulation}
Table~\ref{tab:notation} in the appendix summarizes our notation.
We model execution as a sequence of actions and observations. At step $t$,
the base policy selects $a_t$ from the task $x$, recent history, and a
temporary decision view $c_t$:
\begin{equation}
a_t\sim\pi_\theta(\cdot\mid x,h_t^{\mathrm{recent}},c_t).
\end{equation}
ACG maintains persistent execution memory $G_t$ and constructs $c_t$ under
the remaining context budget $B_t$. The view is read-only and is not itself
a permanent requirement graph.

After executing $a_t$, the resulting observation is recorded with its action
origin and incorporated into memory for the next decision. This closes the
interaction loop: evidence accumulation changes what can be retrieved at
subsequent steps, while the base policy retains control of planning and tool
use. The persistent record and the temporary view therefore serve different
roles: the former preserves execution evidence, and the latter selects what
the policy sees now.
The complete update order is summarized in Algorithm~\ref{alg:acg-loop}.

\subsection{Source-Grounded Execution Memory}
Observations and explicit state records are decomposed into memory units. We
write a unit as $m=(\mathrm{id}_m,z_m,\tau_m,\rho_m,\mathcal O_t(m),\chi_m)$,
where $z_m$ is the content, $\tau_m$ its type, $\rho_m$ its source
references, $\mathcal O_t(m)$ its observed occurrences, and $\chi_m$ its
structured fields such as object identifiers and attributes. An occurrence
records the producing event, action, observation, source, and temporal order.
Repeated content may share a memory unit while preserving each occurrence:
\begin{equation}
\mathcal O_{t+1}(m)=\mathcal O_t(m)\cup\Delta\mathcal O_{t+1}(m).
\end{equation}
Each occurrence retains its event identity and tool-call origin when
available. Source links identify where a record came from; they do not certify
truth or task completion. Reusing a content unit therefore never merges the
execution occurrences that produced it.

As execution proceeds, source-grounded concept anchors support navigation
across events without replacing exact evidence with generated summaries.
Earlier occurrences remain available, and temporal order does not by itself
declare an observation false or obsolete.

\subsection{Relational Organization and Contextual Retrieval}
ACG retrieves with both the original task and the latest interaction. Let
$L_t^x$ and $L_t^e$ be their ranked memory lists. We select unique seeds by
round-robin interleaving:
\begin{equation}
\begin{aligned}
S_t&=\operatorname{Unique}\!\left[L_t^x(1),L_t^e(1),L_t^x(2),L_t^e(2),\ldots\right]_{1:K},\\
\widetilde{\mathcal C}_t&=S_t\cup\{m:\operatorname{dist}_{G_t}(m,S_t)\le d\}
\cup\operatorname{LocalCluster}(S_t).
\end{aligned}
\end{equation}
The graph connects memory units that are mutual nearest neighbors in
embedding space. For hierarchical organization, these relations are combined
with source affinity: units sharing an original source receive additional
weight, normalized by the number of units from that source. A
structural-entropy organizer~\citep{li2016structural} groups this weighted
graph into local clusters (Appendix~\ref{app:reproducibility}). Semantic
links support navigation across related content; source affinity keeps
co-produced evidence accessible together. Neither relation creates a
requirement attribution. The depth $d$ is fixed; adaptation
concerns the selected evidence and its level of detail at each decision.

The task-seeded list supplies global objective cues, whereas the
event-seeded list restores information about the immediate interaction.
Round-robin selection balances the two ranked lists when selecting unique
seeds; graph and concept expansion then supplies local context around each seed.

\subsection{Requirement-Grounded Decision Views}
Requirements are represented by exact instruction spans. Attribution is
tracked at the level of individual action or tool-call occurrences.
Structured origin declarations identify the requirement anchors associated
with a call; their identifiers must resolve to existing requirements. An
observation may inherit these anchors through an unambiguous parent action.
Missing or ambiguous attribution remains unassigned rather than being filled
by semantic similarity. This validates the provenance of an association,
not whether the observation is true or the requirement has been satisfied.

Let
$A_t(\omega)$ denote the requirements supported by the validated origin of
occurrence $\omega$. The source-qualified candidates for requirement $r$ are
\(Q_t(r)=\{m\in\widetilde{\mathcal C}_t:\exists\omega\in\mathcal O_t(m),\ r\in A_t(\omega)\}\).
Let $E_t^{\mathrm{cur}}$ be memory units from the current event. Historical
units are ranked for each requirement, while qualified current-event units
are retained separately:
\begin{equation}
\begin{aligned}
C_t(r)&=\operatorname{TopK}_{m\in Q_t(r)\setminus E_t^{\mathrm{cur}}}s_r(m)\\
&\quad\cup\bigl(Q_t(r)\cap E_t^{\mathrm{cur}}\bigr),\\
\mathcal I_t&=\{(r,m,\omega):m\in C_t(r),\ \omega\in\mathcal O_t(m),\ r\in A_t(\omega)\}.
\end{aligned}
\end{equation}
Here $\operatorname{TopK}$ selects up to eight historical memory units,
and $s_r(m)$ is their embedding similarity to the requirement text.
Ties are resolved by the latest associated event and
then by memory identifier. This ranking applies only after source
qualification and cannot create a provenance link.
The incidences are a temporary projection and are not written back to the
persistent graph; they do not certify task completion.

\subsection{Budget-Aware Context Construction}
The renderer orders membership, current-event, evidence, and cross-reference
blocks. For an ordered block $b_j$, it admits the block only when the
complete rendered view remains within budget:
\begin{equation}
V^{(j)}=\begin{cases}V^{(j-1)}\oplus b_j,&\operatorname{Tok}(\operatorname{Render}(V^{(j-1)}\oplus b_j))\le B_t,\\V^{(j-1)},&\text{otherwise.}\end{cases}
\end{equation}
Under tighter budgets, ACG shortens the representation of
requirement--evidence associations and prioritizes current-event evidence.
Evidence shared by multiple requirements is expanded once and referenced
elsewhere, reducing duplication while preserving its associations. Records
omitted from the view remain available in persistent memory.

\section{Experiments}
\subsection{Experimental Setup}
\label{sec:setup}
\paragraph{Benchmarks.}
We evaluate three benchmarks with fixed task sets within each benchmark:
DeepPlanning~\citep{zhang2026deepplanning}, BrowseComp-Plus~\citep{chen2025browsecompplus},
and SWE-bench Lite~\citep{jimenez2024swebench,swebenchlite}. DeepPlanning
contains constrained shopping and travel planning tasks, BrowseComp-Plus
requires retrieval from a fixed corpus, and SWE-bench Lite evaluates
repository-level software fixes. The fixed sets contain 360, 830, and 300
tasks, respectively. We use each benchmark's native success criterion;
Appendix~\ref{app:reproducibility} describes outcome accounting.

\paragraph{Baselines.}
We compare with ReAct~\citep{yao2023react}, COMPASS~\citep{wan2025compass},
TDP*~\citep{li2026tdp}, CUGA~\citep{cuga}, and ACON~\citep{kang2025acon}.
ReAct interleaves reasoning and actions; COMPASS separates execution from
context management; TDP* maintains dependency-oriented local histories; CUGA
and ACON provide framework-level and history-compression baselines.
TDP* denotes our reproduction of TDP.

\paragraph{Models and Execution Budgets.}
We use GPT-5.6-luna and DeepSeek-v4-flash with a nominal budget of 100 external actions per task
and an executor output limit of 4,096 tokens per call. Throughout the paper,
steps refer to recorded external-action counts; LLM calls are counted
separately and include auxiliary computation. Benchmark tools and evaluators
are fixed within each model setting.

\paragraph{Metrics.}
We report benchmark success rates and task-level inference expenditure,
including auxiliary calls and input plus output tokens. Table~\ref{tab:main-results}
reports standard deviations from 5,000 task-bootstrap resamples, measuring
task-sampling uncertainty rather than variability across repeated runs.

\subsection{Main Results}
Table~\ref{tab:main-results} reports task success rates on the three
benchmarks. Scores use the fixed task sets and outcome-accounting policy
described in Appendix~\ref{app:reproducibility}.

\begin{table}[t]
\centering
\caption{Task success rates (\%). All methods use the same task set within each benchmark. Avg. is the equal-weight arithmetic mean of the three benchmark rates; it is not pooled over tasks. Subscripts after $\pm$ in the benchmark columns are bootstrap standard deviations.}
\label{tab:main-results}
\resizebox{\textwidth}{!}{\begin{tabular}{llrrrr}
\toprule
\multirow{2}{*}{Model} & \multirow{2}{*}{Method} & \multicolumn{3}{c}{Dataset} & \multirow{2}{*}{Avg.} \\
\cmidrule(lr){3-5}
 & & DeepPlanning & BrowseComp-Plus & SWE-bench Lite & \\
\midrule
\multirow{6}{*}{\shortstack[l]{GPT-5.6-\\luna}}
 & ReAct & \underline{\rate{10.8}{1.64}} & \underline{\rate{62.4}{1.71}} & \underline{\rate{60.3}{2.84}} & \underline{44.5} \\
 & COMPASS & \rate{6.4}{1.28} & \rate{48.1}{1.72} & \rate{12.7}{1.93} & 22.4 \\
 & TDP* & \rate{6.9}{1.36} & \rate{10.5}{1.05} & \rate{22.3}{2.39} & 13.3 \\
 & CUGA & \rate{0.0}{0.00} & \rate{10.96}{1.08} & \rate{3.67}{1.08} & 4.88 \\
 & ACON & \rate{7.8}{1.42} & \rate{10.2}{1.04} & \rate{50.0}{2.89} & 22.7 \\
\cmidrule(lr){2-6}
 & \textbf{ACG (Ours)} & \bfrate{12.4}{1.75} & \bfrate{73.5}{1.87} & \bfrate{64.7}{2.76} & \textbf{50.2} \\
\midrule
\multirow{6}{*}{\shortstack[l]{DeepSeek-\\v4-flash}}
 & ReAct & \rate{6.4}{1.28} & \bfrate{68.2}{1.59} & \underline{\rate{56.7}{2.84}} & \underline{43.8} \\
 & COMPASS & \underline{\rate{10.6}{1.62}} & \rate{8.7}{0.99} & \rate{4.0}{1.12} & 7.74 \\
 & TDP* & \rate{2.8}{0.88} & \rate{18.1}{1.34} & \rate{10.7}{1.79} & 10.5 \\
 & CUGA & \rate{0.0}{0.00} & \rate{8.6}{0.98} & \rate{3.7}{1.07} & 4.07 \\
 & ACON & \rate{3.1}{0.91} & \rate{48.4}{1.73} & \rate{31.3}{2.69} & 27.6 \\
\cmidrule(lr){2-6}
 & \textbf{ACG (Ours)} & \bfrate{11.7}{1.70} & \underline{\rate{62.7}{1.64}} & \bfrate{63.3}{2.76} & \textbf{45.9} \\
\bottomrule
\end{tabular}}
\end{table}

ACG achieves the highest average success for both models in
Table~\ref{tab:main-results}. Its gains depend on the benchmark and model: GPT-5.6-luna
ACG reaches 73.5\% on BrowseComp-Plus and 64.7\% on SWE-bench Lite, while
DeepPlanning is 12.4\%. With DeepSeek-v4-flash, ACG reaches 62.7\%,
63.3\%, and 11.7\% on BrowseComp-Plus, SWE-bench Lite, and DeepPlanning,
respectively. The equal-weight averages are 50.2\% for Luna ACG and 45.9\%
for DeepSeek ACG; these
averages summarize the three benchmarks and should not be read as a pooled
estimate. Taken together, these results support a conditional claim:
organizing execution evidence can help on some long-horizon task mixtures,
but the effect is not uniform across models or benchmarks.

The benchmark-level comparison sharpens this picture: ACG ranks first in
five of the six model--benchmark settings and second in the remaining one.
ReAct is the strongest baseline in five settings; DeepSeek DeepPlanning
instead favors COMPASS among the baselines. ACG therefore competes with
different reference methods across tasks, rather than benefiting only from
one weak comparator. Nevertheless, its 12.4\% and 11.7\% DeepPlanning
success rates show that constrained planning remains difficult despite
leading the compared methods.

\subsection{Task Structure and Baseline Performance}
\label{sec:task-structure}
\begin{figure}[t]
\centering
\includegraphics[width=\textwidth]{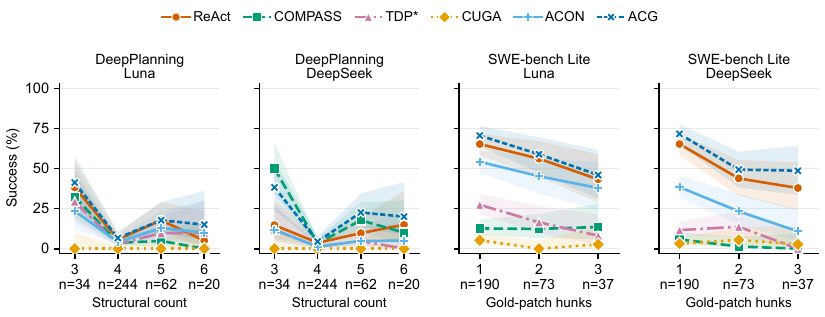}
\caption{Task success conditioned on recorded task structure. Luna and
DeepSeek denote GPT-5.6-luna and DeepSeek-v4-flash. Points
show the success fraction in each group; shaded bands are pointwise Wilson
95\% confidence intervals, and $n$ denotes the number of tasks per group
for each displayed method. DeepPlanning pools shopping subquery counts and
travel hard-constraint counts; equal counts across these families do not
imply equal difficulty. SWE-bench Lite uses reference-patch hunk counts.
Lines connect observed group estimates without fitting.}
\label{fig:task-structure}
\end{figure}

Figure~\ref{fig:task-structure} compares groupwise success on DeepPlanning
and SWE-bench Lite. Structural counts describe task composition, not minimum
execution length; BrowseComp-Plus document counts describe evidence load
and are not included on this scale.

On SWE-bench Lite, ACG exceeds ReAct in all three displayed structural
groups for both models. For DeepSeek, ACG succeeds on 71.6\% of one-hunk
tasks and 48.6\% of three-hunk tasks, compared with ReAct's 65.3\%
and 37.8\%. ACG's advantage therefore extends beyond single-location
repairs, although success still decreases as changes become more dispersed.
The smaller three-hunk group has wider uncertainty, so these point estimates
do not establish a statistically significant interaction with patch size.

Luna COMPASS remains near 12--14\% across the three SWE groups, below
ReAct throughout. A flatter curve therefore need not indicate robustness:
preserving success on simpler tasks matters alongside handling dispersed changes.

DeepPlanning shows a non-monotonic pooled profile. Shopping and travel
contribute different proportions at each count, so this pattern reflects
task composition rather than an isolated within-family complexity effect.
The contrast with SWE cautions against explaining performance across
domains using a single scalar notion of difficulty.

\subsection{Execution Length and Task Outcomes}
\label{sec:execution-length}
\begin{figure}[t]
\centering
\includegraphics[width=\textwidth]{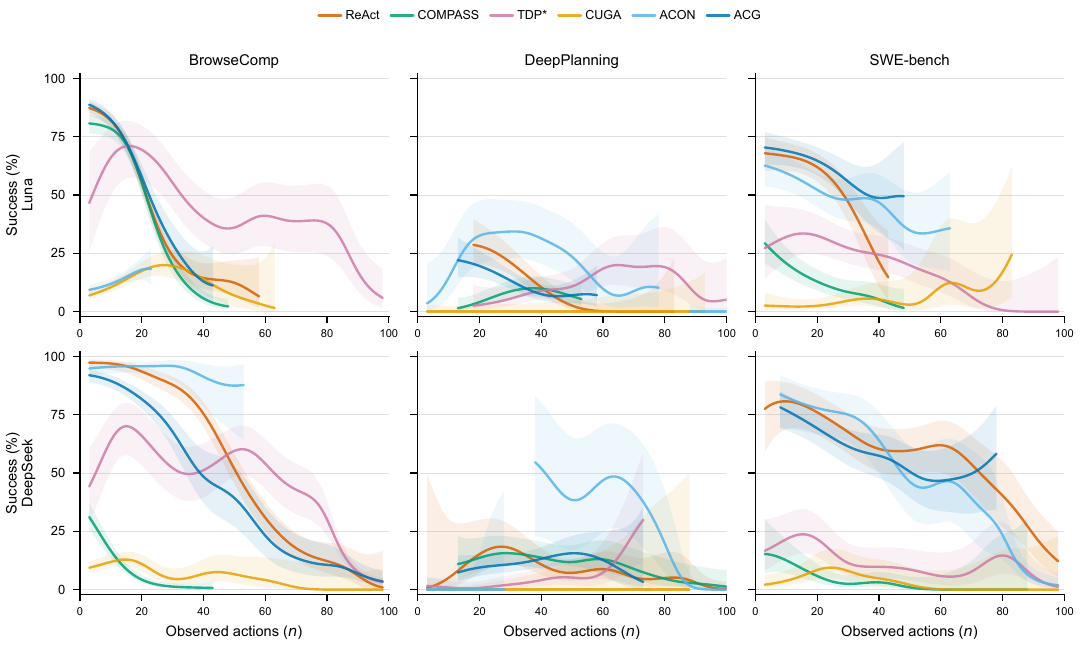}
\caption{Success conditioned on observed execution length, with Luna above
and DeepSeek below. Curves smooth five-action-bin success fractions using
sample-count-weighted Gaussian kernels (bandwidth seven actions).
Shading shows approximate pointwise 95\% intervals. The horizontal axis
measures realized actions, not an imposed budget; smoothing details are in
Appendix~\ref{app:reproducibility}.}
\label{fig:execution-length}
\end{figure}

\paragraph{Long executions often concentrate unsuccessful tasks.}
Figure~\ref{fig:execution-length} complements the structural breakdown with
the length of the realized execution. For DeepSeek ReAct on BrowseComp-Plus,
279 of 287 tasks ending within 1--20 actions succeed (97.2\%), compared with
24 of 170 ending within 61--100 actions (14.1\%). On SWE-bench Lite, the
corresponding fractions are 39/46 (84.8\%) and 34/98 (34.7\%). These raw
group counts support the broad decline in the smoothed curves: longer
observed executions are not simply successful solutions that require more
work. They also contain substantial unsuccessful effort. Execution length
therefore needs to be interpreted jointly with task outcome when evaluating
an agent's use of its action budget.

\paragraph{The pattern is method- and task-dependent.}
ACG exhibits this dependence as well. With DeepSeek on BrowseComp-Plus,
its raw success fraction falls from 88.7\% among executions ending within
1--20 actions to 6.9\% within 61--100 actions. On SWE-bench Lite, ACG
retains 50.0\% success in the latter group, compared with ReAct's 34.7\%.
These are conditional comparisons of different trajectory groups, not
matched-task estimates of a benefit from executing longer.

The decline is not shared uniformly across methods or benchmarks.
DeepPlanning remains difficult even in many shorter-execution groups,
whereas some methods retain relatively high conditional success over parts
of the BrowseComp-Plus range. A high curve in a selected length range does
not imply superior overall performance: the distribution of tasks across
lengths differs by method, and failures can accumulate outside that range.
For example, DeepSeek ACON has high conditional success in several early
BrowseComp-Plus bins while its overall success remains below ReAct in
Table~\ref{tab:main-results}. This motivates assessing whether ACG converts
otherwise unsuccessful execution into completed tasks, rather than rewarding
either longer deliberation or shorter trajectories in isolation.

Observed length reflects both task difficulty and agent behavior; these
associations do not establish that changing the action limit improves success.

\subsection{Model-Call Overhead and Context Cost}
\label{sec:efficiency}
\begin{figure}[t]
\centering
\includegraphics[width=\textwidth]{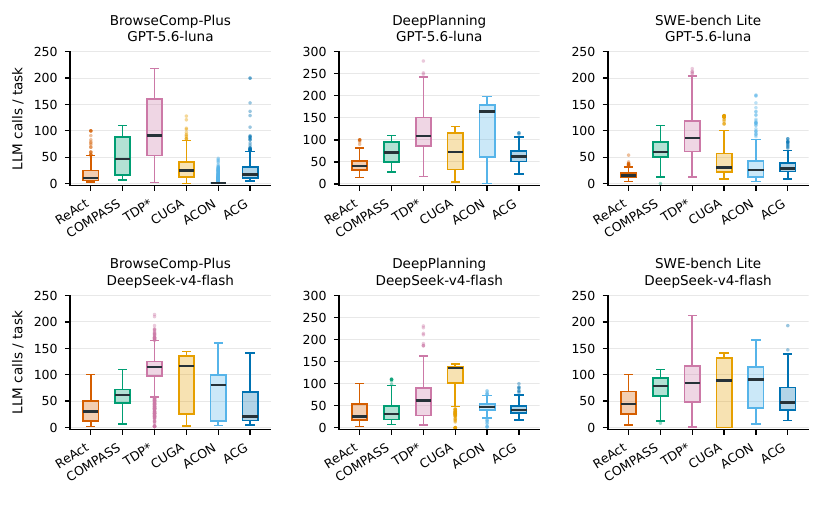}
\caption{Task-level LLM call counts, including successful and unsuccessful
executions and auxiliary calls. Boxes show medians and interquartile ranges;
whiskers extend to 1.5 IQR and dots denote outliers. Vertical scales are
shared between models within each benchmark.}
\label{fig:llm-call-cost}
\end{figure}

Figure~\ref{fig:llm-call-cost} includes executor and auxiliary calls for
successful and unsuccessful tasks. Equal action limits do not imply equal
inference costs, since one external action can involve multiple model calls.

\paragraph{More model calls need not yield higher task success.}
On SWE-bench Lite with GPT-5.6-luna, median call counts are 15 for ReAct,
60 for COMPASS, 86 for TDP*, and 30.5 for CUGA. Yet their success rates are
60.3\%, 12.7\%, 22.3\%, and 3.7\%, respectively
(Table~\ref{tab:main-results}). In particular, TDP* uses approximately
5.7 times as many calls at the median as ReAct, while its median total
token use is 1.075 million versus 0.128 million, approximately 8.4 times
as large. The additional work does not translate into higher success,
and the larger token ratio shows that call counts alone understate the cost difference.

\paragraph{ACG's success gains have different call costs across settings.}
On SWE-bench Lite, Luna ACG uses a median of 29 calls versus 15 for
ReAct, while success rises from 60.3\% to 64.7\%. With DeepSeek, the
corresponding medians are 47 and 44 calls, with success of 63.3\% and
56.7\%, respectively. On BrowseComp-Plus, DeepSeek ACG uses fewer calls
at the median (21 versus 30), but also has lower success (62.7\% versus
68.2\%). Thus ACG does not offer a uniform reduction in inference effort:
its completion gains and call expenditure must be assessed jointly.
These method-level comparisons do not establish that additional calls
cause either success or failure.

\subsection{Execution and Delivery Diagnostics}
Tables~\ref{tab:execution-diagnostics} and~\ref{tab:execution-recovery}
compare ACG with all five baselines using the same diagnostic definitions.
Rates are computed separately within successful and unsuccessful tasks;
the following comparisons concern the unsuccessful group.

\paragraph{ACG's advantage is not a uniform reduction in repetition.}
On SWE-bench Lite with DeepSeek, repeated execution occurs in 8.2\% of
assessable ACG failures, versus 27.9\% for COMPASS, 96.4\% for TDP*,
64.9\% for CUGA, and 24.8\% for ACON. ReAct is lower still at 6.2\%.
With Luna, ACG and ReAct both have 0.0\% repetition, compared with
4.6\%, 84.2\%, 12.8\%, and 12.1\% for those four baselines.
Thus ACG avoids much of the repeated execution seen in several competing
systems, but reduced repetition alone cannot explain its advantage over ReAct.

\paragraph{Progress toward delivery is distinct from correctness.}
For DeepSeek on SWE-bench Lite, missing deliverables account for 0.9\%
of assessable ACG failures, compared with 26.2\% for ReAct, 94.1\% for
COMPASS, 78.4\% for TDP*, 71.3\% for CUGA, and 76.2\% for ACON.
This profile is consistent with ACG sustaining execution through delivery,
while the remaining failures still require better solution quality.
Continuation alone is insufficient: both ACG and ReAct resume accepted
execution after 100.0\% of assessed rejection episodes in this failed-task
group, yet the tasks ultimately fail. These observations locate behavioral
differences between systems without isolating the contribution of graph
retrieval, provenance qualification, or rendering individually.

Taken together, these diagnostics separate avoiding stalled execution from
producing a correct solution. The distinction explains why fewer repeated
actions or more frequent delivery should be interpreted alongside the
official success rates, rather than used as substitutes for them.

\section{Conclusion}
We introduce Adaptive Consistency Graph (ACG), a framework for organizing
and retrieving execution evidence in long-horizon agents. ACG incrementally
constructs a source-grounded graph and assembles requirement-conditioned
context views under a bounded token budget, preserving underlying records
for subsequent retrieval. Baseline analyses show that longer execution and
greater model-call expenditure do not consistently translate into higher
task success, motivating the joint evaluation of completion quality and
inference cost. In the current matched comparisons, ACG improves success on
selected benchmark--model pairs, but the pattern is uneven elsewhere. Taken
together, ACG provides a concrete, testable approach to studying whether
structured access to execution history can improve this trade-off.

\subsection*{AI Use Statement}
Generative AI tools were used to assist with code development, figure preparation, and language polishing. All AI-assisted code, figures, and text were reviewed and verified by the authors, who take full responsibility for the final manuscript.

\subsection*{Reproducibility Statement}
Section~\ref{sec:setup} specifies the benchmarks, comparison methods,
models, and metrics. Appendix~\ref{app:reproducibility} details task
accounting, statistical uncertainty, and context-selection settings;
Appendix~\ref{app:execution-diagnostics} defines the supplementary
execution diagnostics.

\bibliography{iclr2027_conference}
\bibliographystyle{iclr2027_conference}

\clearpage
\appendix
\setcounter{figure}{0}
\setcounter{table}{0}
\renewcommand{\theHfigure}{appendix.\arabic{figure}}
\renewcommand{\theHtable}{appendix.\arabic{table}}

\section{Algorithmic Details}
\label{app:algorithm}
The following pseudocode specifies the online ACG loop independently of any
benchmark-specific tool interface. It makes explicit the separation between
persistent evidence storage, temporary context construction, and base-agent
execution.

\begin{table}[t]
\centering
\caption{Notation used in the ACG formulation.}
\label{tab:notation}
\small
\begin{tabular}{@{}ll@{}}
\toprule
Symbol & Meaning \\
\midrule
$x$ & task instruction \\
$a_t$, $o_t$ & action and resulting observation at step $t$ \\
$h_t^{\mathrm{recent}}$ & recent action--observation history \\
$G_t$ & persistent source-grounded execution memory \\
$m$ & memory unit; $\mathcal O_t(m)$ stores its occurrences \\
$L_t^x$, $L_t^e$ & task-seeded and event-seeded ranked lists \\
$S_t$, $\widetilde{\mathcal C}_t$ & selected seeds and expanded candidate set \\
$C_t(r)$, $\mathcal I_t$ & requirement-qualified candidates and incidences \\
$c_t$, $B_t$ & temporary decision view and remaining token budget \\
$K$, $d$ & seed limit and graph-expansion depth \\
\bottomrule
\end{tabular}
\end{table}

\begin{algorithm}[t]
\caption{Adaptive Consistency Graph decision loop}
\label{alg:acg-loop}
\begin{minipage}{0.94\linewidth}
\small
\textbf{Input:} task $x$, base policy $\pi_\theta$, budget $B_t$, seed limit $K$, depth $d$.\\
\textbf{Initialize:} $G_0\gets\emptyset$ and $h^{\mathrm{recent}}_0\gets\emptyset$.\\
\textbf{For each step $t$ until termination:}
\begin{enumerate}\itemsep0pt
\item Rank task- and event-seeded records, then interleave the top $K$ unique seeds.
\item Expand the persistent graph to depth $d$ and project candidates by task requirement.
\item Render $c_t=\operatorname{Render}(\mathcal I_t,B_t)$ and sample
$a_t\sim\pi_\theta(\cdot\mid x,h^{\mathrm{recent}}_t,c_t)$.
\item If $a_t$ is terminal, return it and $G_t$; otherwise execute it, observe $o_t$,
ingest $(a_t,o_t)$ with its source into $G_{t+1}$, and append it to recent history.
\end{enumerate}
\end{minipage}
\end{algorithm}

The ingest operation preserves event identity, source references, and
requirement-origin information when it is available; it does not infer a
missing attribution from semantic similarity. The renderer may omit or
shorten records to satisfy $B_t$, while omitted records remain in $G_t$.

\section{Evaluation and Reproducibility Details}
\label{app:reproducibility}
\subsection{ACG Context Selection}
\paragraph{Context-selection settings.}
The ACG configuration uses eight unique retrieval seeds and a fixed
one-hop graph expansion. Requirement-specific ranking retains up to eight
historical candidates per requirement; qualified evidence from the current
event is additionally retained as a candidate. The memory view has a
5,000-token cap, further reduced when less context space is available.
Candidate retention does not guarantee full textual expansion: the final
view is subject to the rendering budget described in Section~\ref{sec:method}.

\subsection{Structural Organizer}
\paragraph{Structural-entropy organizer.}
Each mutual-nearest-neighbor pair has unit semantic edge weight. For a
source shared by $n>1$ memory units, each pair receives source-affinity
weight $1/(n-1)$; identical source-member groups are counted once. Semantic
and source-affinity weights are added for hierarchical organization.
Zero-volume groups contribute zero to the entropy, and an edgeless graph
has score zero.

For completeness, the hierarchy organizer uses the standard two-level
structural entropy. For a partition $P$ of graph nodes, let
$d_v$ be the weighted degree of node $v$,
$\operatorname{vol}(C)=\sum_{v\in C}d_v$, and
$\operatorname{vol}(G)=\sum_{v}d_v$. Let $g_C$ be the total weight of edges
with exactly one endpoint in $C$. Its score is
\begin{equation}
H(P)=-\sum_{C\in P}\frac{g_C}{\operatorname{vol}(G)}
\log_2\frac{\operatorname{vol}(C)}{\operatorname{vol}(G)}
-\sum_{C\in P}\sum_{v\in C}\frac{d_v}{\operatorname{vol}(G)}
\log_2\frac{d_v}{\operatorname{vol}(C)}.
\end{equation}
This is an adopted organization criterion, rather than a learned ACG
objective or a guarantee about task success. It supplies local clusters for
retrieval expansion; it does not determine actions or certify evidence.

\subsection{Task Accounting and Uncertainty}
\paragraph{Task sets and outcome accounting.}
Comparisons use the same 360 DeepPlanning, 830 BrowseComp-Plus, and 300
SWE-bench Lite task identifiers within each model setting. The baseline
diagnostics in Appendix~\ref{app:execution-diagnostics} report task-level
rates separately by method and outcome.

\paragraph{Action counts and usage.}
Step-based analyses use the recorded external-action count under the nominal
100-action budget; the execution-length figure displays actions up to 100.
LLM calls are a separate quantity and include recorded auxiliary work,
such as history compression. Total tokens are input plus output tokens;
cached input is a subset of input, not an additional cost term.

\paragraph{Task-bootstrap uncertainty.}
For Figure~\ref{fig:execution-length}, five-action-bin success rates are
smoothed with sample-count-weighted Gaussian kernels of bandwidth seven
actions. A segment requires three consecutive nonempty bins and at least
30 tasks; empty bins are not bridged. Weighted-score intervals use kernel
effective sample sizes and are pointwise, not simultaneous or corrected
for smoothing bias.

Within each available method--model--benchmark combination, we sample
$N$ task outcomes with replacement from its $N$ recorded binary outcomes
and recompute success rate. We repeat this 5,000 times with fixed seeds
derived from the combination identity and seed 20260923. The sample
standard deviation of these rates is reported in percentage points.
This procedure conditions on the observed executions and estimates
task-sampling variation only. In particular, resampling an all-failure
set gives zero standard deviation; this does not establish that the
method has zero success probability on other tasks or repeated executions.
For the table's equal-weight three-benchmark average, we report the
arithmetic mean of the three cell estimates without a standard deviation.

\section{Observable Execution and Delivery Diagnostics}
\label{app:execution-diagnostics}
We compare ACG and the baselines using deterministic observations of
execution traces and final artifacts. We distinguish six overlapping categories: interface
rejection, operation-precondition rejection, explicit budget exhaustion,
exact repeated execution, missing deliverables, and delivered artifacts
that fail official evaluation. These observations describe execution and
delivery, rather than an exhaustive or causal taxonomy of task failure.

\paragraph{Operational definitions.}
Interface rejection requires an explicit invalid-call, argument, or tool
schema response. Operation-precondition rejection requires a recognized
tool-specific refusal, such as a missing target object or an unmet
submission prerequisite; internal implementation exceptions are not
included. Budget exhaustion requires an explicit terminal budget reason,
not merely a long trajectory. Repetition means two consecutive calls with
identical tool names and arguments, or two consecutive copies of a
two- or three-call pattern with distinct calls within the pattern.
Only JSON key order is normalized. Rejected interface calls break the
sequence, and unobserved calls prevent an absence claim. Repetition can
be legitimate, including repeated inspection after a state change.

\paragraph{Deliverables and evaluation.}
Deliverable presence is determined from a task-owned final answer on
BrowseComp-Plus, a final report or shopping cart on DeepPlanning, and a
final prediction patch on SWE-bench Lite. Nonempty artifacts establish
presence, not correctness. An explicitly empty artifact establishes
absence; a missing file alone does not. Delivered-but-failed requires both
artifact presence and a completed official evaluation reporting failure.
When artifact absence is established, delivered-but-failed is false by definition.

\paragraph{Rates and continuation.}
Each category uses its own assessable-task denominator, separately for
successful and unsuccessful tasks.
An obstacle episode begins with an interface or precondition rejection
and ends at the next accepted external execution, or at the end of the
trace. Episodes do not overlap. We count subsequent attempts including
the resuming call, so immediate resumption has delay one. Episodes ending
without resumption distinguish further rejected attempts from no further
attempt. Resumption means accepted execution, not repaired task state or
eventual success. Task-level percentages use assessable tasks within the
same method, benchmark, model, and outcome group. Continuation percentages
use obstacle episodes as their denominator; repeated episodes within a task
are not independent task samples.

\begingroup
\footnotesize
\setlength{\tabcolsep}{2.5pt}
\renewcommand{\arraystretch}{1.10}
\begin{longtable}{@{}llrrrrrr@{}}
\caption{Task-level execution and delivery observations, reported as percentages; F/S denotes final failure/success. Each category cell is positive / assessable $\times 100$ within its outcome group, using that category's assessable-task denominator. Categories overlap and are not mutually exclusive, so percentages are not additive; a dash denotes no denominator.}\label{tab:execution-diagnostics}\\
\toprule
Method & Outcome & Interface & Condition & Budget & Repeat & No output & Eval. fail \\
\midrule
\endfirsthead
\multicolumn{8}{l}{Table~\thetable{} continued}\\
\toprule
Method & Outcome & Interface & Condition & Budget & Repeat & No output & Eval. fail \\
\midrule
\endhead
\midrule
\multicolumn{8}{r}{Continued on next page}\\
\endfoot
\bottomrule
\endlastfoot
\multicolumn{8}{l}{\textbf{BrowseComp-Plus / GPT-5.6-luna}}\\*
ReAct & F & 2.6\% & 0.0\% & 1.4\% & 0.0\% & 7.7\% & 92.3\% \\*
 & S & 0.8\% & 0.0\% & 0.0\% & 0.0\% & 0.0\% & 0.0\% \\
COMPASS & F & 87.5\% & 0.0\% & 47.0\% & 67.5\% & 75.4\% & 24.6\% \\*
 & S & 55.4\% & 0.0\% & 0.0\% & 15.3\% & 0.0\% & 0.0\% \\
TDP* & F & 96.8\% & 0.0\% & 0.0\% & 43.9\% & 72.8\% & 26.6\% \\*
 & S & 96.6\% & 0.0\% & 0.0\% & 51.7\% & 0.0\% & 0.0\% \\
CUGA & F & 26.4\% & 0.5\% & 0.0\% & 2.1\% & 22.7\% & 75.0\% \\*
 & S & 65.9\% & 0.0\% & 0.0\% & 5.5\% & 0.0\% & 0.0\% \\
ACON & F & 0.3\% & 0.0\% & 0.0\% & 0.0\% & 94.1\% & 5.9\% \\*
 & S & 0.0\% & 0.0\% & 0.0\% & 0.0\% & 0.0\% & 0.0\% \\
ACG & F & 1.3\% & 0.0\% & 1.3\% & 0.0\% & 1.3\% & 98.7\% \\*
 & S & 0.0\% & 0.0\% & 0.0\% & 0.0\% & 0.0\% & 0.0\% \\
\addlinespace
\multicolumn{8}{l}{\textbf{BrowseComp-Plus / DeepSeek-v4-flash}}\\*
ReAct & F & 72.3\% & 0.0\% & 3.8\% & 0.0\% & 3.8\% & 96.2\% \\*
 & S & 82.5\% & 0.0\% & 0.0\% & 0.0\% & 0.0\% & 0.0\% \\
COMPASS & F & 95.8\% & 0.0\% & 1.1\% & 1.6\% & 88.5\% & 11.5\% \\*
 & S & 59.7\% & 0.0\% & 0.0\% & 0.0\% & 0.0\% & 0.0\% \\
TDP* & F & 87.6\% & 0.0\% & 0.0\% & 13.4\% & 92.4\% & 7.6\% \\*
 & S & 79.3\% & 0.0\% & 0.0\% & 10.0\% & 0.0\% & 0.0\% \\
CUGA & F & 11.2\% & 26.9\% & 5.3\% & 31.6\% & 74.2\% & 25.8\% \\*
 & S & 50.7\% & 0.0\% & 0.0\% & 32.4\% & 0.0\% & 0.0\% \\
ACON & F & 3.0\% & 0.0\% & 95.3\% & 1.2\% & 95.3\% & 4.7\% \\*
 & S & 17.4\% & 0.0\% & 0.0\% & 0.2\% & 4.7\% & 0.0\% \\
ACG & F & 0.0\% & 0.0\% & 33.2\% & 0.3\% & 33.2\% & 66.8\% \\*
 & S & 0.2\% & 0.0\% & 0.0\% & 0.0\% & 0.0\% & 0.0\% \\
\addlinespace
\multicolumn{8}{l}{\textbf{DeepPlanning / GPT-5.6-luna}}\\*
ReAct & F & 3.6\% & 5.4\% & 1.4\% & 6.8\% & 0.0\% & 100.0\% \\*
 & S & 5.4\% & 0.0\% & 0.0\% & 2.9\% & 0.0\% & 0.0\% \\
COMPASS & F & 5.6\% & 7.8\% & 18.6\% & 6.2\% & 12.5\% & 86.9\% \\*
 & S & 0.0\% & 0.0\% & 0.0\% & 0.0\% & 0.0\% & 0.0\% \\
TDP* & F & 79.0\% & 22.1\% & 5.9\% & 69.1\% & 6.2\% & 92.3\% \\*
 & S & 100.0\% & 14.3\% & 4.0\% & 90.5\% & 0.0\% & 0.0\% \\
CUGA & F & 88.5\% & 3.4\% & 0.0\% & 39.3\% & 35.8\% & 64.2\% \\*
 & S & -- & -- & -- & -- & -- & -- \\
ACON & F & 0.0\% & 12.1\% & 76.4\% & 69.7\% & 24.8\% & 74.5\% \\*
 & S & 0.0\% & 21.7\% & 0.0\% & 69.6\% & 0.0\% & 0.0\% \\
ACG & F & 0.0\% & 1.6\% & 0.0\% & 30.6\% & 4.5\% & 94.7\% \\*
 & S & 0.0\% & 2.9\% & 0.0\% & 31.4\% & 0.0\% & 0.0\% \\
\addlinespace
\multicolumn{8}{l}{\textbf{DeepPlanning / DeepSeek-v4-flash}}\\*
ReAct & F & 76.5\% & 0.0\% & 50.5\% & 19.3\% & 5.1\% & 89.9\% \\*
 & S & 50.0\% & 0.0\% & 0.0\% & 0.0\% & 0.0\% & 0.0\% \\
COMPASS & F & 11.5\% & 2.2\% & 13.1\% & 6.5\% & 7.2\% & 92.7\% \\*
 & S & 6.5\% & 5.3\% & 0.0\% & 5.3\% & 0.0\% & 0.0\% \\
TDP* & F & 64.6\% & 8.5\% & 0.7\% & 57.3\% & 2.9\% & 96.2\% \\*
 & S & 80.0\% & 12.5\% & 0.0\% & 50.0\% & 0.0\% & 0.0\% \\
CUGA & F & 88.4\% & 22.5\% & 0.0\% & 25.0\% & 95.8\% & 4.2\% \\*
 & S & -- & -- & -- & -- & -- & -- \\
ACON & F & 0.0\% & 0.0\% & 92.7\% & 25.0\% & 63.8\% & 20.2\% \\*
 & S & 9.1\% & 0.0\% & 0.0\% & 18.2\% & 0.0\% & 0.0\% \\
ACG & F & 13.0\% & 0.0\% & 0.0\% & 6.9\% & 2.5\% & 97.4\% \\*
 & S & 9.5\% & 0.0\% & 0.0\% & 2.6\% & 0.0\% & 0.0\% \\
\addlinespace
\multicolumn{8}{l}{\textbf{SWE-bench Lite / GPT-5.6-luna}}\\*
ReAct & F & 89.1\% & 0.8\% & 0.0\% & 0.0\% & 0.8\% & 99.1\% \\*
 & S & 91.7\% & 0.6\% & 0.0\% & 0.0\% & 0.0\% & 0.0\% \\
COMPASS & F & 97.7\% & 1.9\% & 30.2\% & 4.6\% & 77.1\% & 12.9\% \\*
 & S & 89.5\% & 0.0\% & 0.0\% & 0.0\% & 0.0\% & 0.0\% \\
TDP* & F & 99.6\% & 2.7\% & 0.0\% & 84.2\% & 49.4\% & 31.1\% \\*
 & S & 97.0\% & 1.5\% & 0.0\% & 80.6\% & 0.0\% & 0.0\% \\
CUGA & F & 89.3\% & 35.1\% & 3.2\% & 12.8\% & 57.8\% & 39.1\% \\*
 & S & 81.8\% & 54.5\% & 0.0\% & 18.2\% & 0.0\% & 0.0\% \\
ACON & F & 28.0\% & 0.7\% & 4.8\% & 12.1\% & 20.0\% & 80.0\% \\*
 & S & 26.0\% & 1.3\% & 0.0\% & 3.3\% & 0.0\% & 0.0\% \\
ACG & F & 67.9\% & 1.9\% & 0.0\% & 0.0\% & 0.0\% & 100.0\% \\*
 & S & 51.0\% & 3.6\% & 0.0\% & 0.0\% & 0.0\% & 0.0\% \\
\addlinespace
\multicolumn{8}{l}{\textbf{SWE-bench Lite / DeepSeek-v4-flash}}\\*
ReAct & F & 80.0\% & 12.4\% & 25.6\% & 6.2\% & 26.2\% & 73.6\% \\*
 & S & 78.8\% & 6.0\% & 0.0\% & 1.8\% & 0.0\% & 0.0\% \\
COMPASS & F & 87.8\% & 6.3\% & 69.6\% & 27.9\% & 94.1\% & 5.9\% \\*
 & S & 58.3\% & 0.0\% & 0.0\% & 8.3\% & 0.0\% & 0.0\% \\
TDP* & F & 32.5\% & 11.4\% & 0.0\% & 96.4\% & 78.4\% & 21.6\% \\*
 & S & 40.6\% & 12.5\% & 0.0\% & 87.5\% & 0.0\% & 0.0\% \\
CUGA & F & 56.2\% & 25.4\% & 5.7\% & 64.9\% & 71.3\% & 13.1\% \\*
 & S & 63.6\% & 9.1\% & 0.0\% & 27.3\% & 0.0\% & 0.0\% \\
ACON & F & 41.3\% & 6.3\% & 76.0\% & 24.8\% & 76.2\% & 20.3\% \\*
 & S & 30.9\% & 5.3\% & 0.0\% & 18.1\% & 0.0\% & 0.0\% \\
ACG & F & 1.8\% & 8.2\% & 0.9\% & 8.2\% & 0.9\% & 94.7\% \\*
 & S & 2.1\% & 10.6\% & 0.0\% & 6.3\% & 0.0\% & 0.0\% \\
\addlinespace
\end{longtable}
\endgroup

\begingroup
\footnotesize
\setlength{\tabcolsep}{3pt}
\renewcommand{\arraystretch}{1.10}
\begin{longtable}{@{}llrrrr@{}}
\caption{Continuation after an interface or operation-precondition rejection, reported as percentages; F/S denotes final failure/success. Resumed, Reject-only, and No further attempt use their total observed episodes as the denominator; Delay is the median subsequent interface-attempt count until resumption. These are event-level observations, and continuation does not imply task success; a dash denotes no denominator or unavailable metric.}\label{tab:execution-recovery}\\
\toprule
Method & Outcome & Resumed & Reject-only & No further attempt & Delay \\
\midrule
\endfirsthead
\multicolumn{6}{l}{Table~\thetable{} continued}\\
\toprule
Method & Outcome & Resumed & Reject-only & No further attempt & Delay \\
\midrule
\endhead
\midrule
\multicolumn{6}{r}{Continued on next page}\\
\endfoot
\bottomrule
\endlastfoot
\multicolumn{6}{l}{\textbf{BrowseComp-Plus / GPT-5.6-luna}}\\*
ReAct & F & 100.0\% & 0.0\% & 0.0\% & 1 \\*
 & S & 100.0\% & 0.0\% & 0.0\% & 1 \\
COMPASS & F & 95.8\% & 0.4\% & 3.8\% & 1 \\*
 & S & 100.0\% & 0.0\% & 0.0\% & 1 \\
TDP* & F & 98.1\% & 0.8\% & 1.1\% & 1 \\*
 & S & 100.0\% & 0.0\% & 0.0\% & 1 \\
CUGA & F & 96.0\% & 0.7\% & 3.3\% & 1 \\*
 & S & 100.0\% & 0.0\% & 0.0\% & 1 \\
ACON & F & 50.0\% & 0.0\% & 50.0\% & 1 \\*
 & S & -- & -- & -- & -- \\
ACG & F & 100.0\% & 0.0\% & 0.0\% & 1 \\*
 & S & -- & -- & -- & -- \\
\addlinespace
\multicolumn{6}{l}{\textbf{BrowseComp-Plus / DeepSeek-v4-flash}}\\*
ReAct & F & 100.0\% & 0.0\% & 0.0\% & 1 \\*
 & S & 100.0\% & 0.0\% & 0.0\% & 1 \\
COMPASS & F & 61.5\% & 36.6\% & 1.9\% & 4 \\*
 & S & 100.0\% & 0.0\% & 0.0\% & 3 \\
TDP* & F & 96.9\% & 1.2\% & 1.9\% & 1 \\*
 & S & 100.0\% & 0.0\% & 0.0\% & 1 \\
CUGA & F & 29.2\% & 1.6\% & 69.2\% & 1 \\*
 & S & 100.0\% & 0.0\% & 0.0\% & 1 \\
ACON & F & 100.0\% & 0.0\% & 0.0\% & 1 \\*
 & S & 100.0\% & 0.0\% & 0.0\% & 1 \\
ACG & F & -- & -- & -- & -- \\*
 & S & 100.0\% & 0.0\% & 0.0\% & 1 \\
\addlinespace
\multicolumn{6}{l}{\textbf{DeepPlanning / GPT-5.6-luna}}\\*
ReAct & F & 100.0\% & 0.0\% & 0.0\% & 1 \\*
 & S & 100.0\% & 0.0\% & 0.0\% & 1 \\
COMPASS & F & 100.0\% & 0.0\% & 0.0\% & 1 \\*
 & S & -- & -- & -- & -- \\
TDP* & F & 100.0\% & 0.0\% & 0.0\% & 1 \\*
 & S & 100.0\% & 0.0\% & 0.0\% & 1 \\
CUGA & F & 95.0\% & 3.3\% & 1.7\% & 1 \\*
 & S & -- & -- & -- & -- \\
ACON & F & 100.0\% & 0.0\% & 0.0\% & 2.5 \\*
 & S & 100.0\% & 0.0\% & 0.0\% & 1 \\
ACG & F & 100.0\% & 0.0\% & 0.0\% & 1 \\*
 & S & 100.0\% & 0.0\% & 0.0\% & 1 \\
\addlinespace
\multicolumn{6}{l}{\textbf{DeepPlanning / DeepSeek-v4-flash}}\\*
ReAct & F & 100.0\% & 0.0\% & 0.0\% & 1 \\*
 & S & 100.0\% & 0.0\% & 0.0\% & 1 \\
COMPASS & F & 100.0\% & 0.0\% & 0.0\% & 1 \\*
 & S & 100.0\% & 0.0\% & 0.0\% & 3 \\
TDP* & F & 100.0\% & 0.0\% & 0.0\% & 1 \\*
 & S & 100.0\% & 0.0\% & 0.0\% & 1 \\
CUGA & F & 90.3\% & 5.7\% & 4.0\% & 1 \\*
 & S & -- & -- & -- & -- \\
ACON & F & -- & -- & -- & -- \\*
 & S & 100.0\% & 0.0\% & 0.0\% & 1 \\
ACG & F & 100.0\% & 0.0\% & 0.0\% & 1 \\*
 & S & 100.0\% & 0.0\% & 0.0\% & 1 \\
\addlinespace
\multicolumn{6}{l}{\textbf{SWE-bench Lite / GPT-5.6-luna}}\\*
ReAct & F & 100.0\% & 0.0\% & 0.0\% & 1 \\*
 & S & 100.0\% & 0.0\% & 0.0\% & 1 \\
COMPASS & F & 75.5\% & 22.8\% & 1.7\% & 1 \\*
 & S & 100.0\% & 0.0\% & 0.0\% & 1 \\
TDP* & F & 98.0\% & 1.9\% & 0.1\% & 1 \\*
 & S & 100.0\% & 0.0\% & 0.0\% & 1 \\
CUGA & F & 99.8\% & 0.2\% & 0.0\% & 2 \\*
 & S & 100.0\% & 0.0\% & 0.0\% & 2.5 \\
ACON & F & 100.0\% & 0.0\% & 0.0\% & 1 \\*
 & S & 100.0\% & 0.0\% & 0.0\% & 1 \\
ACG & F & 100.0\% & 0.0\% & 0.0\% & 1 \\*
 & S & 100.0\% & 0.0\% & 0.0\% & 1 \\
\addlinespace
\multicolumn{6}{l}{\textbf{SWE-bench Lite / DeepSeek-v4-flash}}\\*
ReAct & F & 100.0\% & 0.0\% & 0.0\% & 1 \\*
 & S & 100.0\% & 0.0\% & 0.0\% & 1 \\
COMPASS & F & 69.6\% & 26.9\% & 3.4\% & 2 \\*
 & S & 100.0\% & 0.0\% & 0.0\% & 2 \\
TDP* & F & 96.3\% & 0.0\% & 3.7\% & 1 \\*
 & S & 100.0\% & 0.0\% & 0.0\% & 1 \\
CUGA & F & 99.7\% & 0.3\% & 0.0\% & 2 \\*
 & S & 100.0\% & 0.0\% & 0.0\% & 1.5 \\
ACON & F & 98.5\% & 0.0\% & 1.5\% & 1 \\*
 & S & 100.0\% & 0.0\% & 0.0\% & 1 \\
ACG & F & 100.0\% & 0.0\% & 0.0\% & 1 \\*
 & S & 100.0\% & 0.0\% & 0.0\% & 1 \\
\addlinespace
\end{longtable}
\endgroup

\section{Limitations}
\label{app:limitations}
\paragraph{Scope and interpretation.}
ACG organizes available execution evidence; provenance does not establish
that an observation is correct or that a task requirement is satisfied.
Missing or ambiguous action origins can reduce requirement-specific
coverage, and finite context budgets can exclude relevant evidence.
Graph construction and retrieval also introduce computation beyond the
base executor. Our baseline analyses are observational: task structure and
realized execution length are associated with outcomes but do not identify
failure causes. Generalization
beyond the evaluated models, tools, and task sets remains unestablished.

\end{document}